# Convolutional Neural Networks for Multi-class Histopathology Image Classification

Muhammed Talo[a*]

[a] Department of Computer Engineering, Munzur University, Tunceli, Turkey

**Abstract**

There is a strong need for automated systems to improve diagnostic quality and reduce the analysis time in histopathology image processing. Automated detection and classification of pathological tissue characteristics with computer-aided diagnostic systems are a critical step in the early diagnosis and treatment of diseases. Once a pathology image is scanned by a microscope and loaded onto a computer, it can be used for automated detection and classification of diseases. In this study, the DenseNet-161 and ResNet-50 pre-trained CNN models have been used to classify digital histopathology patches into the corresponding whole slide images via transfer learning technique. The proposed pre-trained models were tested on grayscale and color histopathology images. The DenseNet-161 pre-trained model achieved a classification accuracy of 97.89% using grayscale images and the ResNet-50 model obtained the accuracy of 98.87% for color images. The proposed pre-trained models outperform state-of-the-art methods in all performance metrics to classify digital pathology patches into 24 categories.

# 1. Introduction

Pathologists manually examine tissue samples using a microscope to determine if a disease is present in a tissue. The diagnosis of pathological diseases has been performed by analyzing various tissue samples which usually obtained by a biopsy procedure [1]. The extracted tissue sections are stained to reveal its high-level structure and then placed on a glass slide in order to analyze the morphological structures [2].

In recent years, the microscopes are being replaced by digital scanners. So, the entire histology slide is scanned by a digital whole slide scanner and saved as a whole slide image (WSI). The analysis of tissue characteristics from WSIs has opened a new era in digital pathology [3]. Computer-aided diagnostic (CAD) systems can perform the diagnosis and classification of the diseases using WSI images. However, the diagnosis of digital pathology images is a tedious task because of the large image size of the WSIs. The high-resolution WSIs are usually divided into partial patches and the analyses are performed individually on each sample.

In the analysis of histological images, different interpretations can be seen among different pathologists [4]. Since the WSIs are very large in size, it is not easy for experts to diagnose disease types by looking at all patches. The computer-aided diagnosis systems can overcome these problems. Hence, there is a strong demand for the development of computer-assisted diagnostic tools that can reduce the workload of pathologists by helping them for a fast and precise diagnosis [5].

The convolutional neural networks (CNNs) is become a method of choice in medical image analysis. In recent years, deep learning architectures, especially CNNs are the most successful type of methods has been applied to medical image analysis tasks such as detection [6-9] and classification [10-11]. The CNN architectures have an end-to-end structure, which learns high-level representations from the data [12]. Deep learning algorithms have also achieved tremendous success in pathology image processing, such as mitosis detection [13], tissue grading (classification) [14], and nuclei segmentation [15] from the high-resolution images. In histopathology image analysis, the color and texture based features of histology images are used for the segmentation and classification tasks [16].

Several studies have been conducted for automated classification and segmentation of histopathology images using deep learning methods. Saha et al. [17], implemented a deep learning based model with handcrafted features (textual, morphological and intensity features) for automated detection of mitoses from breast histopathology WSIs. The authors reported a 92% precision and 88% recall for the proposed architecture. Han et al. [18] have implemented a structured deep learning model for multi-classification of breast cancer from histopathology images. They achieved an average of 93.2% classification accuracy to classify subclasses of breast cancer such as lobular carcinoma, ductal carcinoma, and fibroadenoma. Zheng et al. [19], developed a convolutional neural network to classify breast lesions. The authors used a two class (benign and malignant) breast tumor dataset and reached a classification accuracy of 96.6%. They also used a 15-class breast tumors dataset and reported 96.4% average classification accuracy. Jia et al. [20] proposed a fully connected network using multiple instance learning algorithm to segment cancerous regions in histopathological images. Komura et al. [21], presents a concise review of studies that use machine learning approaches for digital histopathology image analysis.

The contribution of this study is fourfold: (1) An end-to-end pre-trained models were developed for automated classification of histopathology images. (2) The color and grayscale images are classified into 24 categories using all available data in Kimia Path24 dataset. (3) The classification performance of different pre-trained models was compared for both color and grayscale images. (4) The proposed pre-trained models demonstrate a better classification accuracy than the existing studies on the literature.

The remainder of this paper is organized as follows. The materials and methods are presented in the next section. The experiments and results are reported in Section 3 and discussions are given in Section 4. Finally, Section 6 presents the contributions and future works.

## 2. Material and Methods

### 2.1. Dataset

The Kimia Path24 histopathology dataset used in this study was released by Babaie et al. [22], which contains 24 whole-slide tissue images. The dataset is publicly available and shows various body parts with different texture patterns. The images in the Kimia Path24 dataset were obtained by TissueScope LE 1.0. The dataset contains a total of 23,916 images and 1325 of these images selected for testing. All images have an equal size of 1000×1000 pixels. The selected test patches were extracted from the WSIs and their locations were stained to white on the scans. Thus, the use of test patches for the training set was prevented.

The previous studies conducted on the Kimia Path24 dataset were converted the color images to grayscale and the classification performance was given on grayscale images. In this study, the classification of images is performed using both color and grayscale images in order to use the color and texture features. Few sample grayscale and color images from Kimia Path24 are shown in Fig.1.

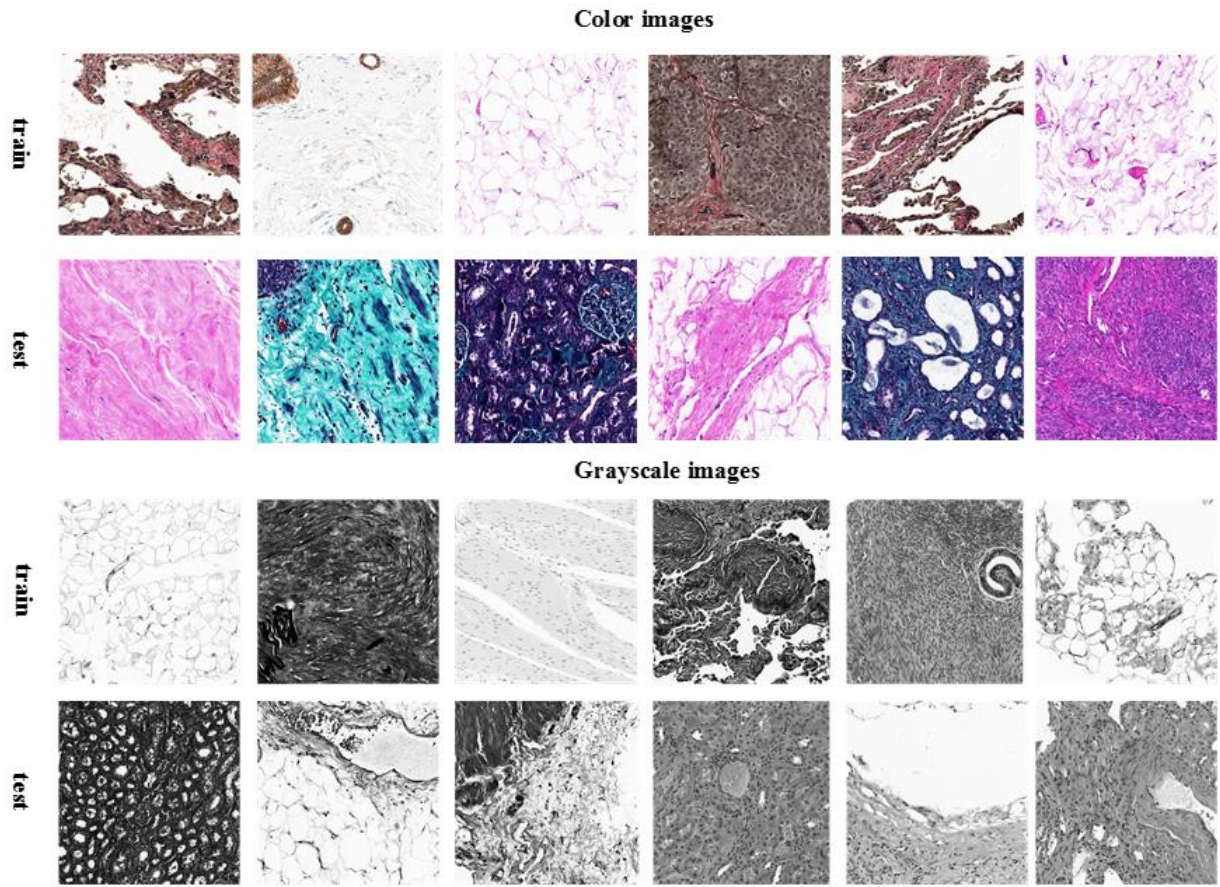

**Figure 1.** A few sample color and grayscale images from Kimia Path24 dataset. Each row shows the training and test images of color and grayscale images.

## 2.2. Deep transfer learning

Convolutional neural networks automatically learn the best representative features from the raw data instead of using traditional machine learning techniques that benefit from the handcrafted features. A typical convolutional neural network consists of a series of building blocks such as convolutional layers, activation layers, pooling layers, and fully-connected layers. The shallow CNN architectures were constructed stacking a couple of the building blocks together like AlexNet [23] and VggNet [24]. However, the modern CNN architectures are deeper as compared to the shallow CNN architectures and use progressively more complex connections among alternating layers, such as ResNet [25] and DenseNet [26].

In this study, the ResNet-50 and DenseNet-161 pre-trained CNN architectures are employed to classify digitalized pathology images into 24 different classes. ResNet and DenseNet architectures have been used in the construction of new state-of-the-art models. Both of these architectures trained on a subset of ImageNet database which contains 1.2 million images belongs to 1000 classes.

**ResNet:** The Residual neural networks (ResNet) was developed by He et al. [25], got the first place in the ImageNet Large Scale Visual Recognition Challenge (ILSVRC 2015) with 3.57% error rate. The authors used a 152 layer deep CNN architecture in the computation. The ResNet architecture popularized the idea of using deeper networks as compared to AlexNet, which has eight layers, and VggNet with up to 19 layers. The ResNet architecture introduced skip connections, also known as residual connections to avoid information loss during training of deep network. Skip connection technique enables to train very deep networks and can boost the performance of the model. The authors were able to train a 1001-layer deep model using residual connections. The ResNet architecture mainly composed of residual blocks. In shallow neural networks, consecutive hidden layers are linked to each other, however, in the ResNet architecture, there are also connections among residual blocks. The main compelling advantages of residual connections in ResNet architecture; the connections preserve the gained knowledge during training and speed up the training time of the model by increasing the capacity of the network. A block diagram of the pre-trained ResNet-50 model used in this study is shown in Fig.2.

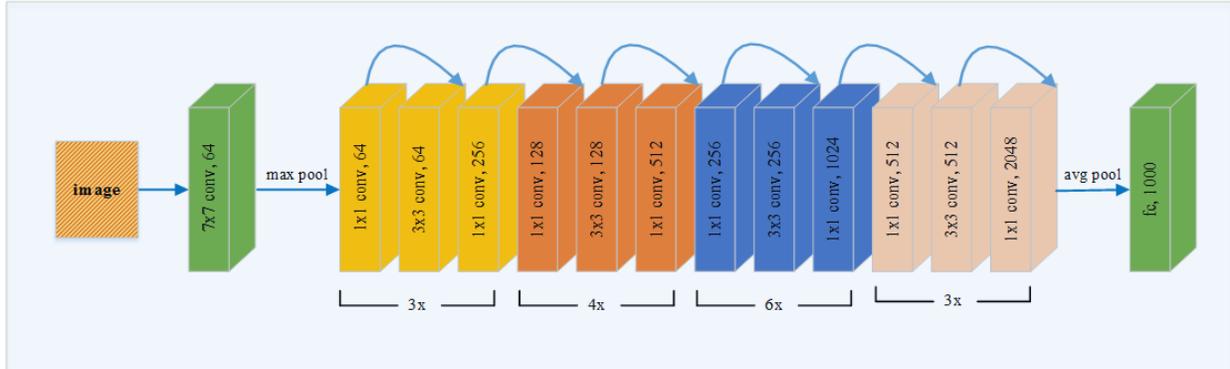

**Figure 2.** A block diagram representation of pre-trained Resnet-50 architecture.

**DenseNet:** Dense convolutional networks (DenseNet) developed by Huang, Liu and Maaten [26] had the best classification performance on publicly available image datasets such as CIFAR-10 and ImageNet in 2017. The DenseNet architecture uses dense connections in the construction of the model, other than direct connections among the hidden layers of the network, as in the ResNet architecture. DenseNet was built with a structure that connects each layer to later layers. Hence, important features learned by any layers of the network is shared within the network. In other words, the extra links among the layers of DenseNet boost the information flow through the whole network. In this way, the training of the deep network becomes more efficient and the performance of the model increases. The DenseNet architecture uses fewer parameters than similar CNN architectures for the training of the network because of the direct transmission of the feature maps to all subsequent layers. Additionally, using dense connections reduce the overfitting of the model on tasks which have smaller datasets. A block diagram of DenseNet-161 having four dense blocks is shown in Figure 3.

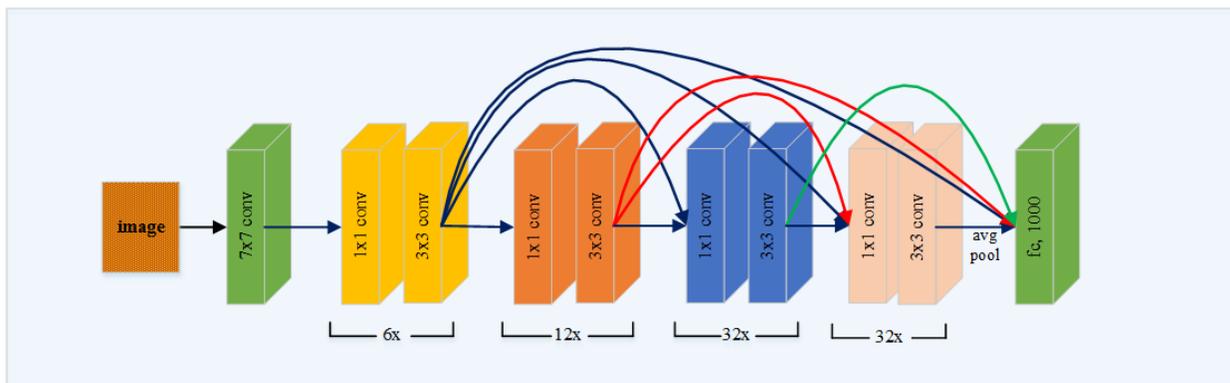

**Figure 3.** The schematic representation of the DenseNet-161 architecture.

Training deep architectures with millions of parameters can take weeks using random weights initialization technique. A large amount of data and powerful computer hardware (GPUs) are required in order to train CNN from starch. The transfer learning technique is commonly used to elevate these problems. With this technique, a CNN model is trained on a huge dataset then the features learned from this model transfer to a model at hand. In the transfer learning technique, the fully connected layer of the model is removed and the remaining layers of the architecture are used as a feature extractor for the new task. Therefore, only the fully connected layers of the new model are trained in the current model. The schematic representation of transfer learning technique used in this study to classify histopathology images is given in Fig.4.

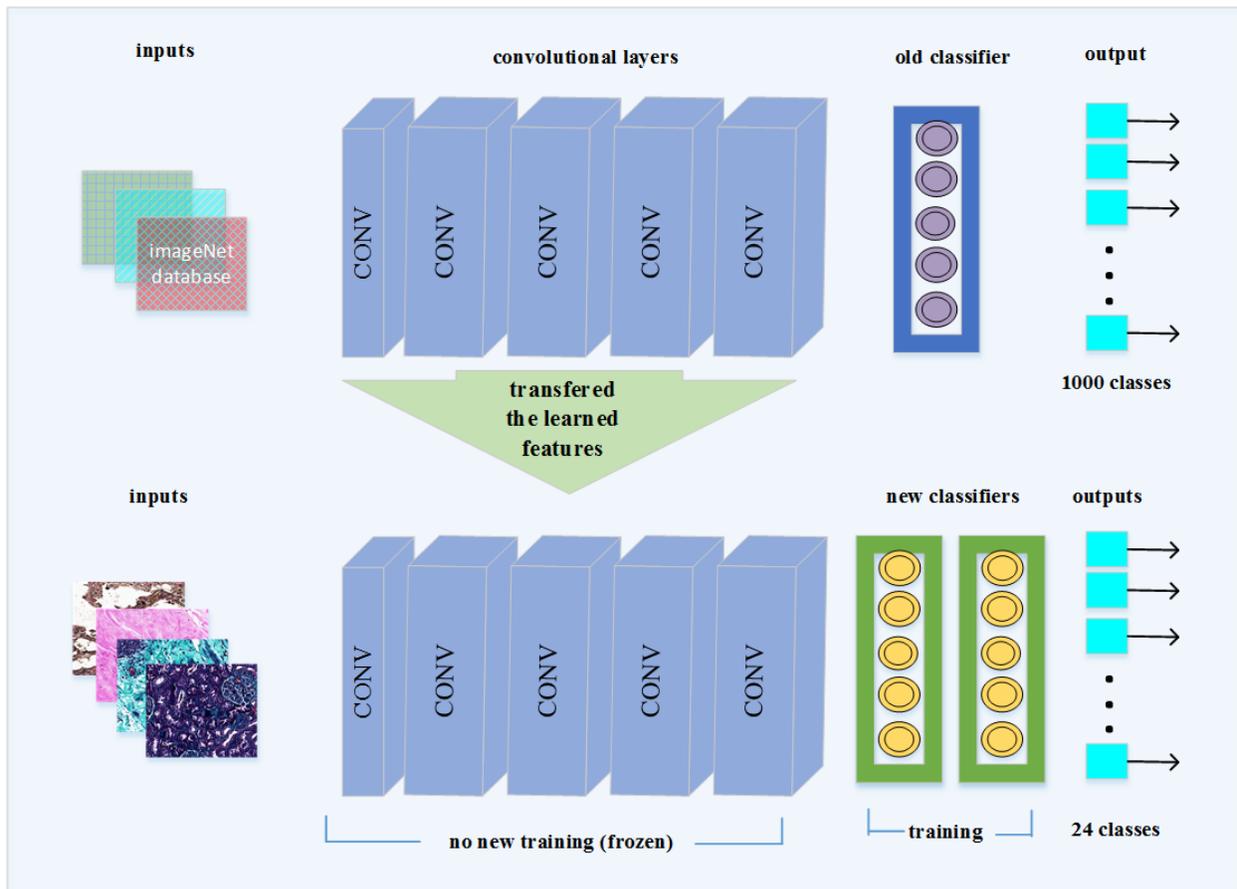

**Figure 4.** The transfer learning method used to classify histopathology images.

## 3. Experiments and Results

The Resnet-50 and DenseNet-161 pre-trained CNN architectures are employed to classify histopathology images into 24 classes. The official Kimia Path24 dataset consists of a total of 23,916 images for training and 1325 images for testing. The training set of 23,916 images are divided into 80% for training and 20% for the validation set in order to have a separate validation set. The performances of the proposed pre-trained CNN models are evaluated on the test set.

PyTorch [27] library and Fastai [28] framework used to implement the proposed pre-trained models. Both of these libraries are open source and build on Python library for machine learning tasks. The training and testing of the ResNet-50 and DenseNet-161 models were performed on NVIDIA GeForce GTX 1080 TI with 11 GB graphics memory.

### 3.1 Experimental Setups

In this study, the transfer learning technique is applied to the ResNet-50 and DensNet-161 pre-trained CNN architectures. The fully connected layer of the DenseNet-161 and ResNet-50 were removed and the convolutional layers of these pre-trained models were used as a *base network* in the new architectures. Two sets of batch normalization, dropout, and fully connected layers are respectively added to the base network. The addition of two batch normalization layers is due to the rapid training of the pre-trained model. The dropout layers are attached to the base network to alleviate the overfitting problem. The overfitting problem causes the model to fail to generalize on the unseen test dataset. Further, in the last layer of proposed models, the Softmax activation function is used to classify the digital pathology images into 24 classes. The overall framework of the customized CNN architecture is given in Fig.5.

Each of the pre-trained models was trained independently on color and grayscale datasets. All the models were trained for the same number of epochs. The training of CNN was performed only in the newly attached layers of the network. With this way, the computation cost of the network in the training process is decreased as compared to the training of the whole network.

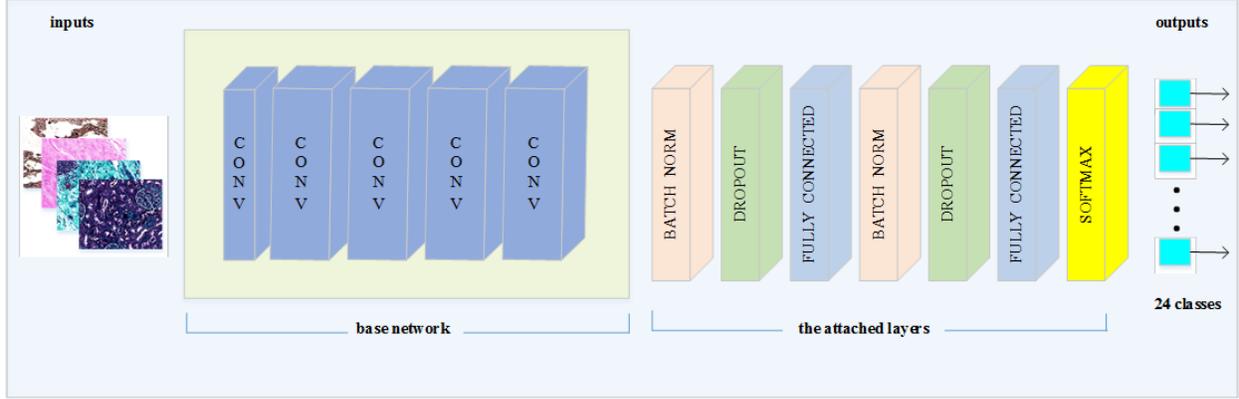

**Figure 5.** An illustration of the customized CNN architecture. The first five feature extractor convolutional blocks which are shown inside the yellow box are used as the base of the network. The rest of the layers are added to the base network.

The stochastic gradient descent (SGD) with momentum, namely RMSprop is employed to optimize the parameters of the network during training. The learning rate value is randomly chosen as 1e-3. The dropout ratios are selected as 0.25 and 0.50, respectively, to regularize the deep models. The batch normalization values were selected as 0.1 for momentum and 1e-5 for epsilon.

The accuracy calculation protocol established by [22] is followed in this study to compare the results of this work with other studies conducted on the Kimia Path24 dataset. There are a total of 1325 test patches, $n_{tot} = 1325$ from 24 distinct sets. Let $R$ represent the set of retrieved images for an experiment and let $\Gamma_s = \{P_s^i | s \in S, \ i = 1, 2, \dots \}$, where $s = 0, 1, \dots, 23$, the *patch-to-scan accuracy*, $\eta_p$ is defined as:

$$\eta_p = \frac{1}{n_{tot}} \sum_s |R \cap \Gamma_s|$$

and the *whole-scan accuracy*, $\eta_w$, is described as follows:

$$\eta_w = \frac{1}{24} \sum_s |R \cap \Gamma_s|$$

The *total accuracy* is defined as:

$$\eta_{total} = \eta_p \times \eta_w$$

using both whole-scan and patch-to-scan accuracies [22]. The Python code for the accuracy calculations is publicly available and can be obtained from the Kimia Lab website [29].

## 3.2 Results

The attached layers of the ResNet-50 and DenseNet-161 models were trained in an end-to-end fashion to classify histopathology images for 50 epochs. Each pre-trained model was trained separately on both color and grayscale images. For each grayscale and color datasets, the training accuracy and the training and validation loss graphs of pre-trained ResNet-50 and DenseNet-161 are shown in Fig.5.

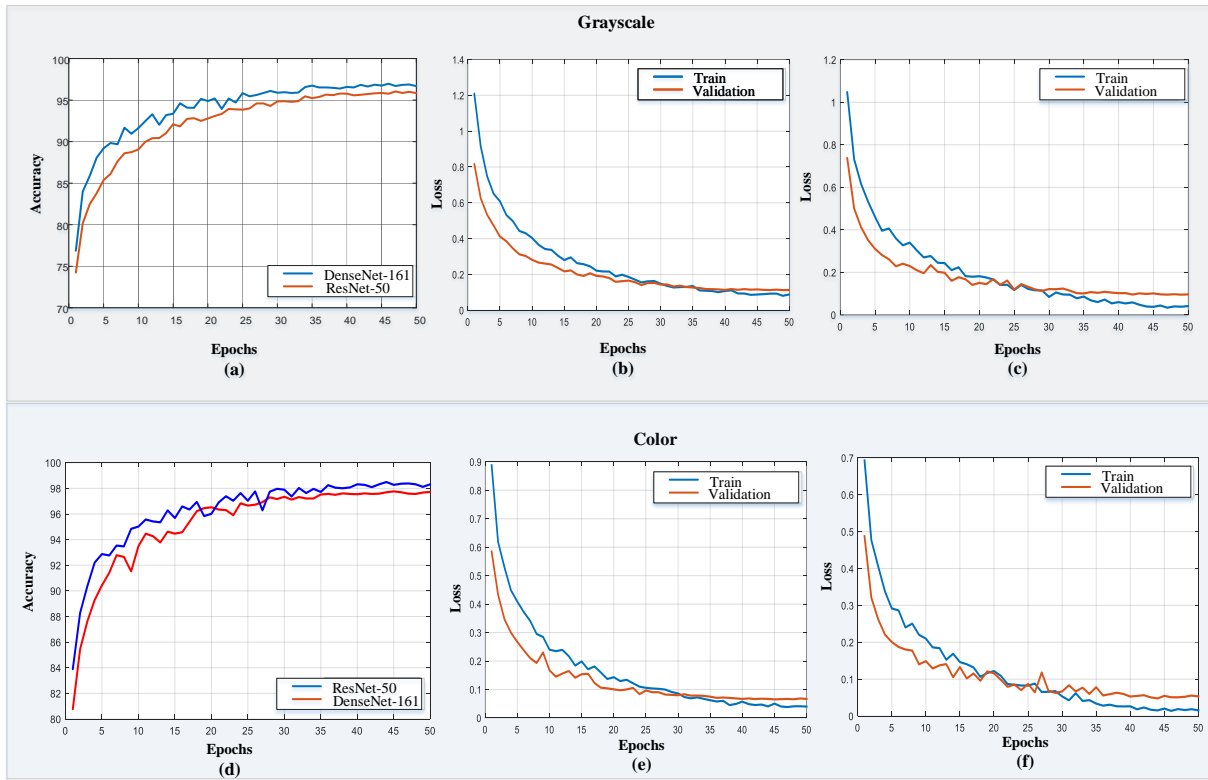

**Figure 5.** The performance of DenseNet-161 and ResNet-50 architectures on grayscale and color datasets: (a) training accuracies of DenseNet-161 and ResNet-50 on grayscale images, (b) training and validation loss of ResNet-50 for grayscale images, (c) training and validation loss of DenseNet-161 for grayscale images, (d) training accuracies of ResNet-50 and DenseNet-161 on color dataset, (e) training and validation loss of ResNet-50 for color images, (f) training and validation loss of DenseNet-161 for color images.

During the training, DenseNet-161 outperforms the ResNet-50 for color images. However, for grayscale images, ResNet-50 obtained better validation accuracy than the DenseNet-161. Table 1 shows the patch-to-scan ($\eta_p$), whole scan ($\eta_w$), and total ($\eta_{total}$) accuracies on test data and the total training time of each pre-trained CNN models for the color and grayscale images.

**Table 1.** Classification results achieved on the test set using ResNet-50 and DenseNet-161 pre-trained models.

| Models | Image format | Training Time (hour : min : sec) | $\eta_p$ (%) | $\eta_w$ (%) | $\eta_{total}$ (%) |
|---|---|---|---|---|---|
| **DenseNet-161** | **grayscale** | **4:14:37** | **97.89** | **97.86** | **95.79** |
| ResNet-50 | grayscale | 3:01:49 | 96.08 | 96.38 | 92.60 |
| DenseNet-161 | RGB | 4:23:50 | 98.64 | 98.63 | 97.28 |
| **ResNet-50** | **RGB** | **4:03:34** | **98.87** | **98.89** | **97.77** |

It can be seen from the Table-2 that; the DenseNet-161 achieves better classification accuracy than the ResNet-50 architecture in the classification of grayscale images on the test set. DenseNet-161 architecture has developed the idea of shortcut connection among the layers of the network that also used by ResNet-50 architecture. DenseNet-161 architecture has more layers than ResNet50. However, surprisingly, in the grayscale images, ResNet-50 pre-trained model outperformed the DensNet-161 in the classification task. The confusion matrixes of DenseNet-161 and ResNet-50 pre-trained models obtained on grayscale and color test datasets (1325 images) are shown in Fig.6.

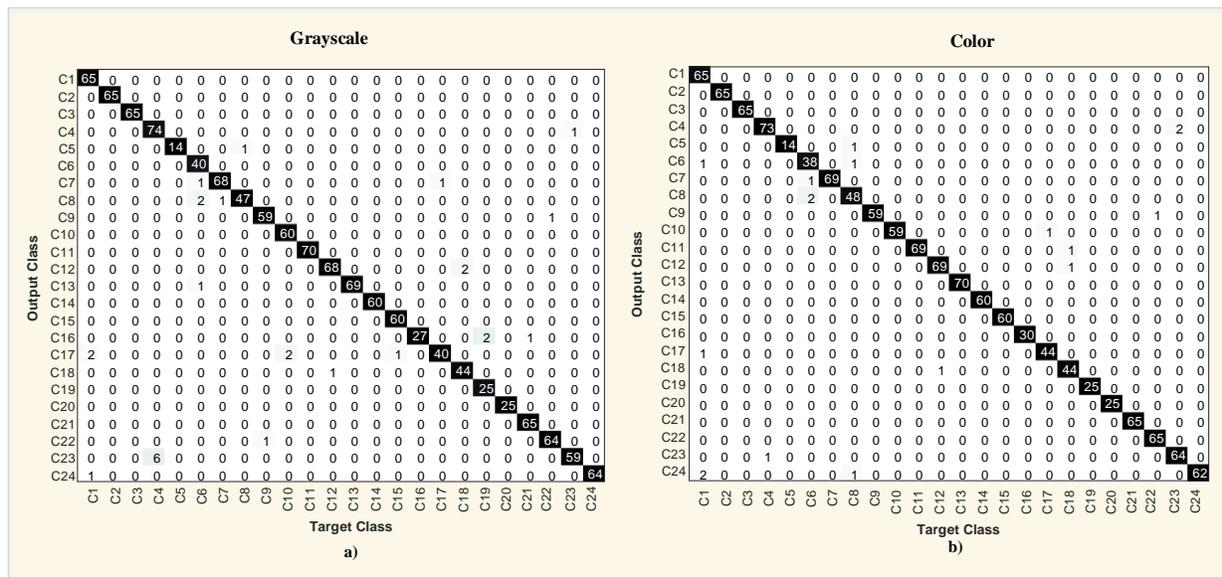

**Figure 6.** The confusion matrixes are obtained using grayscale and color test datasets: (a) DenseNet-161, (b) ResNet-50.

Out of 1325 images, the DenseNet-161 model misclassified 25 images on grayscale test dataset and the ResNet-50 misclassified 18 histopathology images on the color test dataset. The precision, sensitivity, specificity, and F1-Score values of DenseNet-161 model on the grayscale dataset is given in Table 2 for the detailed performance analysis.

Table 2. The classification report for DensNet-161 model using the grayscale test dataset.

| Classes | Precision (%) | Recall (%) | F1-score (%) | Amount of Data |
| --- | --- | --- | --- | --- |
| Class 1 | 0.96 | 1.00 | 0.98 | 65 |
| Class 2 | **1.00** | **1.00** | **1.00** | 65 |
| Class 3 | **1.00** | **1.00** | **1.00** | 65 |
| Class 4 | 0.93 | 0.99 | 0.95 | 75 |
| Class 5 | 1.00 | 0.93 | 0.97 | 15 |
| Class 6 | 0.91 | 1.00 | 0.95 | 40 |
| Class 7 | 0.99 | 0.97 | 0.98 | 70 |
| Class 8 | 0.98 | 0.94 | 0.96 | 50 |
| Class 9 | 0.98 | 0.98 | 0.98 | 60 |
| Class 10 | 0.97 | 1.00 | 0.98 | 60 |
| Class 11 | **1.00** | **1.00** | **1.00** | 70 |
| Class 12 | 0.99 | 0.97 | 0.98 | 70 |
| Class 13 | 1.00 | 0.99 | 0.99 | 70 |
| Class 14 | **1.00** | **1.00** | **1.00** | 60 |
| Class 15 | 0.98 | 1.00 | 0.99 | 60 |
| Class 16 | 1.00 | 0.90 | 0.95 | 30 |
| Class 17 | 0.98 | 0.89 | 0.93 | 45 |
| Class 18 | 0.96 | 0.98 | 0.97 | 45 |
| Class 19 | 0.93 | 1.00 | 0.96 | 25 |
| Class 20 | **1.00** | **1.00** | **1.00** | 25 |
| Class 21 | 0.98 | 1.00 | 0.99 | 65 |
| Class 22 | 0.98 | 0.98 | 0.98 | 65 |
| Class 23 | 0.98 | 0.91 | 0.94 | 65 |
| Class 24 | 1.00 | 0.98 | 0.99 | 65 |

The average precision, recall, and F1-score value for DenseNet-161 is 98% on the grayscale images and 99% for ResNet-50 on the color images. The analyses show that the performance of pre-trained models on the color dataset is better than the grayscale. This may be due to the reason that the ResNet-50 pre-trained model learned both color and textual features during training. However, DenseNet-161 model only learned the textual representations on grayscale data.

## 4. Discussion

Different CNN models, pre-trained networks and feature extraction methods were used to classify histopathology images. Table 3 displays the comparison of the state-of-art studies applied to the Kimia Path24 dataset.

Table 3. Accuracy comparison of this work with other studies that used the same dataset

| Papers | Year | Image Format | Method | $\eta_p$ (%) | $\eta_w$ (%) | $\eta_{total}$ (%) |
|---|---|---|---|---|---|---|
| Babaie et al. [22] | 2017 | Grayscale | BoW | 64.98 | 61.02 | 39.65 |
| Babaie et al. [22] | 2017 | Grayscale | LBP | 66.11 | 62.52 | 41.33 |
| Babaie et al. [22] | 2017 | Grayscale | CNN | 64.98 | 64.75 | 42.07 |
| Kieffer et al. [30] | 2017 | Grayscale | VGG-16 | 65.21 | 64.96 | 42.36 |
| Kieffer et al. [30] | 2017 | Grayscale | Inception-v3 | 74.87 | 76.10 | 56.98 |
| Zhu et al. [31] | 2018 | Grayscale | BoW | 88.07 | 84.50 | 74.41 |
| Zhu et al. [31] | 2018 | Grayscale | MBoW | 89.21 | 85.30 | 76.09 |
| **The proposed** | **2019** | **Grayscale** | **DenseNet-161** | **97.89** | **97.86** | **95.79** |
| **The proposed** | **2019** | **RGB** | **ResNet-50** | **98.87** | **98.89** | **97.77** |

In 2017, Babaie et al. [22], employed the bag of words (BoW), local binary pattern (LBP) histograms, and convolutional neural networks (CNN) approaches to classify grayscale images. The total accuracy value ($\eta_{total}$) for BoW, LBP and CNN approaches were reported as 39.65%, 41.33%, and 42.07%, respectively. The highest classification accuracy, 42.07% was achieved with a CNN model that used 40,513 patches for the training of the network. The same year, Kieffer et al. [30], used data augmentation and transfer learning techniques to train VGG-16 and Inception-v3 architectures. The Inception-v3 pre-trained CNN model achieved the highest total accuracy of 56.98%. In 2018, Zhu et al. [31] proposed a BoW and a variant of BoW with multiple dictionaries (MBoW) approaches to classify histopathology images. The authors excluded patches that has a bright background and used the rest of 25,390 images for the training set. The Bow and MBoW methods reached the total accuracy of 74.41% and 76.09%, respectively.

It can be seen from Table 4 that the proposed DenseNet-161 pre-trained CNN model outperforms other methods in all performance metrics for the classification of grayscale images. While the DenseNet-161 model has achieved a total accuracy of 95.79% for grayscale images, the ResNet-50 model obtains a total accuracy of 97.77% in color images.

The main advantage of this study is that the proposed pre-trained CNN models have a better classification accuracy than other studies in the literature on the Kimia Path24 data set. In addition, the proposed method does not include any hand-made feature extraction technique, i.e., it has a fully automatic end-to-end structure.

**5. Conclusion**

The detection and classification of diseases from tissue samples are performed by pathologists manually. The diagnostic process by an expert using a microscope is time-consuming and expensive. Hence, the automated analysis of tissue samples from histopathology images has critical importance in terms of early diagnosis and treatment. It can also improve the quality of diagnoses and assist the pathologist for the decision making about the critical cases. In this study, the DenseNet-161 and ResNet-50 pre-trained models are employed to classify digital histopathology images into 24 categories. The classification performance of pre-trained models is compared for the color and grayscale images. The proposed DenseNet-161 model tested on grayscale images and obtained the best classification accuracy of 97.89% which outperforms the state-of-art methods. Additionally, the proposed ResNet-50 pre-trained model is tested on color images of Kimia Path24 dataset and achieved a classification accuracy of 98.87%. For future work, an ensemble learning method could be used to increase the classification performance.

**Conflict of interest**

There is no conflict of interest.